\documentclass{article}
\usepackage{microtype}
\usepackage{times}
\usepackage{helvet}
\usepackage{courier}
\usepackage{pdfpages}
\usepackage{graphicx}
\usepackage{algorithm}
\usepackage{algorithmic}
\usepackage{float}
\usepackage{caption}
\usepackage{subcaption}
\usepackage{url}
\usepackage{amsmath, amssymb, mathrsfs}
\usepackage{setspace}

\usepackage{hyperref}

\usepackage[accepted]{icml2019}

\begin{document}
%
\twocolumn[
\icmltitle{Compiler-Level Matrix Multiplication Optimization for Deep Learning}




\begin{icmlauthorlist}
\icmlauthor{Huaqing Zhang}{petuum}
\icmlauthor{Xiaolin Cheng}{huawei}
\icmlauthor{Hui Zang}{huawei}
\icmlauthor{Dae Hoon Park}{amazon} \\
\thanks{petuum}{Petuum, Inc, Sunnyvale, CA, USA} \\
\thanks{huawei}{Huawei Research, Santa Clara, CA, USA} \\
\thanks{amazon}{Amazon, Seattle, WA, USA}
\end{icmlauthorlist}


\icmlkeywords{Machine Learning, ICML}

\vskip 0.3in
]



\begin{abstract}
An important linear algebra routine, GEneral Matrix Multiplication (GEMM), is a fundamental operator in deep learning. Compilers need to translate these routines into low-level code optimized for specific hardware. Compiler-level optimization of GEMM has significant performance impact on training and executing deep learning models. However, most deep learning frameworks rely on hardware-specific operator libraries in which GEMM optimization has been mostly achieved by manual tuning, which restricts the performance on different target hardware. In this paper, we propose two novel algorithms for GEMM optimization based on the TVM framework, a lightweight Greedy Best First Search (G-BFS) method based on heuristic search, and a Neighborhood Actor Advantage Critic (N-A2C) method based on reinforcement learning. Experimental results show significant performance improvement of the proposed methods, in both the optimality of the solution and the cost of search in terms of time and fraction of the search space explored. Specifically, the proposed methods achieve {\bf 24\%} and {\bf 40\%} savings in GEMM computation time over state-of-the-art XGBoost and RNN methods, respectively, while exploring only 0.1\% of the search space. The proposed approaches have potential to be applied to other operator-level optimizations.
\end{abstract}

\section{Introduction}\label{sec_introduction}
In recent years, deep learning has been attracting increasing attention from both academia and industry~\cite{lecun2015deep}. With its own advances in algorithmic and architectural design, significant improvement in computational hardware (e.g. GPU and TPU), and availability of enormous amount of labeled data, deep learning has shown success in many domains, such as computer vision, natural language processing, speech recognition, health care, finance, etc. There is no doubt that deep learning solutions will be increasingly popular in the upcoming years. 



However, computing deep learning models (both training and inference) efficiently on various hardware is a difficult task, which involves end-to-end compiler optimization at several levels from computational graphs to operators, and down to executable code on target hardware~\cite{Chen18}. A computational graph is a global view of operators and data flow among them. Within a graph, operators specify various computation that is required for individual operations on tensors. Current optimization techniques at the computational graph level are hardware agnostic and independent from the implementation of operators within a graph. For example, a standard procedure, operator fusion, combines multiple operators into a single kernel, avoiding latency introduced by write-back and loading of intermediate results into memory. At the operator level, existing optimization is mostly limited to specific implementation for a framework (e.g. TensorFlow XLA) or proprietary library for a particular target device (e.g. Nvidia cuDNN). Such libraries have been built mostly based on expert knowledge and manual tuning to perform effectively and efficiently. 

Deep learning models are expected to run on diverse hardware platforms (CPU, GPU, FPGA, ASIC, SoC, etc.) with very different characteristics that can be leveraged to optimize various deep learning operations. To efficiently map deep learning operation/workload to broader range of hardware, TVM has been proposed~\cite{Chen18} as a general compiler framework for automated tensor operator optimization. In this framework, a configuration space can be defined for each operator. Optimization of an operator is to find a configuration that can optimize a performance metric (e.g., the lowest running time). Configuration is the detailed specification of how an operator is computed and executed on target hardware. For example, in matrix multiplication, tiling is required to make computation more efficient, but various tiling strategies may generate configurations that have significantly different performance. Configuration space for individual operators may have different structures and properties. 

In this paper, we aim at more efficient operator optimization. 
We focus on the GEneral Matrix Multiplication (GEMM) operator which is the fundamental operator in deep learning. GEMM computes the product (multiplication) of two matrices and the operator can be translated into nested loops controlled by a fixed set of parameters. Its optimization can be reduced to finding the optimal combination of parameter values in this set. We analyze structure of its configuration space and design improved tuning approaches for GEMM optimization. The contributions of the paper are three-fold:
\begin{itemize}
\item We consider the relation between different configurations and define the neighbor states of each configuration. We employ a Markov Decision Process (MDP) for exploration over the configuration space.   
\item Based on the neighboring relations within the configuration space, we propose a Greedy Best-First-Search (G-BFS) guided method and a Neighborhood Actor Advantage Critic (N-A2C) method to search for an optimal configuration given two matrices to be multiplied. 
\item We evaluate the performance of our proposed methods in TVM framework for Nvidia Titan XP GPU. We compare them with state-of-the-art GEMM tuners using XGboost~\cite{chen2016xgboost} and RNN controller~\cite{Chen18}. We demonstrate that both our methods can discover high-performance configurations efficiently with smaller fraction of explored configuration space and less time to search. By exploring only 0.1\% of the configuration space, our methods discover configurations of {\bf 24\%} less cost than what the XGBoost method can find and configurations of {\bf 40\%} less cost than what the RNN method can find, for multiplying two $1024 \times 1024$ matrices. 
\end{itemize}

The rest of this paper is organized as follows. We show related work in Section~\ref{sec_works} and describe the GEMM problem in Section~\ref{sec_problem}. In Section~\ref{sec_method}, we propose G-BFS and N-A2C methods and demonstrate experiment results in Section~\ref{sec_simulation}. Section \ref{sec_conclusion} concludes the work.

\section{Related Work}\label{sec_works}


With multiple AI frameworks and a wide range of hardware involved in deep learning applications nowadays, it is important yet challenging for compiler-level optimization to efficiently and flexibly harmonize AI algorithms with the underlying hardware and optimize the performance of various deep learning applications. A large body of work has explored this space and achieved good performance. CuDNN \cite{chetlur2014cudnn} provides highly efficient implementations of various deep neural network layers and is considered the standard for accelerating deep learning on Nvidia GPUs. NNPACK \cite{dukhan2016nnpack} and PCL-DNN \cite{das2016distributed} similarly provide accelerations in x86 processors. Latte \cite{truong2016latte} provides a natural abstraction for specifying new layers and applies domain-specific and general computation graph optimization. XLA (Accelerated Linear Algebra) \cite{leary2017xla} optimizes TensorFlow computations in terms of speed, memory usage, and portability via just-in-time (JIT) compilation or ahead-of-time (AOT) compilation.

However, the aforementioned approaches perform either graph-level or operator-level optimization during compilation and they are not generic enough to accommodate all AI frameworks and hardware. Based on the structure of Halide \cite{ragan2013halide}, \cite{Chen18} proposes a general end-to-end compilation optimization framework combining Neural Network Virtual Machine (NNVM) \cite{nnvm2017} for computation graph optimization and Tensor Virtual Machine (TVM) \cite{Chen18} for tensor operator optimization. 
Currently in TVM, a tuning method based on XGBoost \cite{chen2016xgboost} is considered state-of-the-art method for GEMM configuration optimization and has shown superior performance over other methods \cite{Chen18}, \cite{Tianqi2018}. 

For configuration optimization, the intuitive method is grid search, where all possible configuration candidates are tested sequentially to find the configuration with best performance. It guarantees to find the global optimal configurations, but the number of tested configuration candidates grows exponentially with the dimension of configuration space \cite{bellman2015adaptive}. Therefore, its usage is limited in problems with small search space or in combination with manual search \cite{hinton2012practical,lecun2012efficient,larochelle2007empirical}. Random search is proposed where the configurations are randomly selected to be tested, and is shown empirically and theoretically to be more efficient than grid search for configuration tuning \cite{bergstra2012random}.

As an instance of Bayesian optimization, sequential model-based optimization (SMBO) shows its strength in configuration tuning by iteratively updating the underlying expected improvement, exploring new data through an acquisition function, and training a regression model \cite{hutter2011sequential,bergstra2011algorithms,hoffman2014modular}. The general method has been widely adopted and implemented  \cite{kandasamy2018neural,snoek2012practical}.


From another perspective, a series of evolutionary approaches have been explored, including the broad class of genetic algorithms (GA) \cite{Holland1975,Goldberg1989}, differential evolution \cite{Storn1997}, estimation of distribution algorithms \cite{larranaga2001estimation,bosman2007adapted}, and particle swarm optimization \cite{kennedy2001swarm}.  Evolutionary strategies (ES) \cite{rechenberg1994evolutionsstrategie,schwefel1977numerische} have shown to perform efficiently in configuration tuning. Based on the concept, Covariance Matrix Adaptation Evolution Strategy (CMA-ES), first proposed by \cite{hansen2001completely}, has shown excellent performance by smartly update the mean, step size and covariance matrix for each evolution \cite{LoshchilovH16} . Natural Evolution Strategies \cite{wierstra2014natural} applies natural gradients to update configuration search policy so as to achieve high expected fitness and discover the high-performance configuration.


Recently, researchers at Google apply deep reinforcement learning with a RNN controller to optimize the configurations of neural network architectures and its components \cite{bello2017neural,mirhoseini2017device,ramachandran2018searching,zoph2016neural,pham2018efficient}. The excellent tuning performance in wide range of applications shows the potential of deep reinforcement learning in the configuration tuning area.


\section{Problem Description}\label{sec_problem}
In this section, we describe the concepts of matrix multiplication, GEMM, and matrix tiling. We formulate the GEMM tiling optimization problem at the end of the section.
\subsection{Matrix Multiplication}

\begin{figure}[t]
    \vspace{0.5cm}
    \centering
    \begin{subfigure}[b]{0.22\textwidth}
    \includegraphics[width=\linewidth]{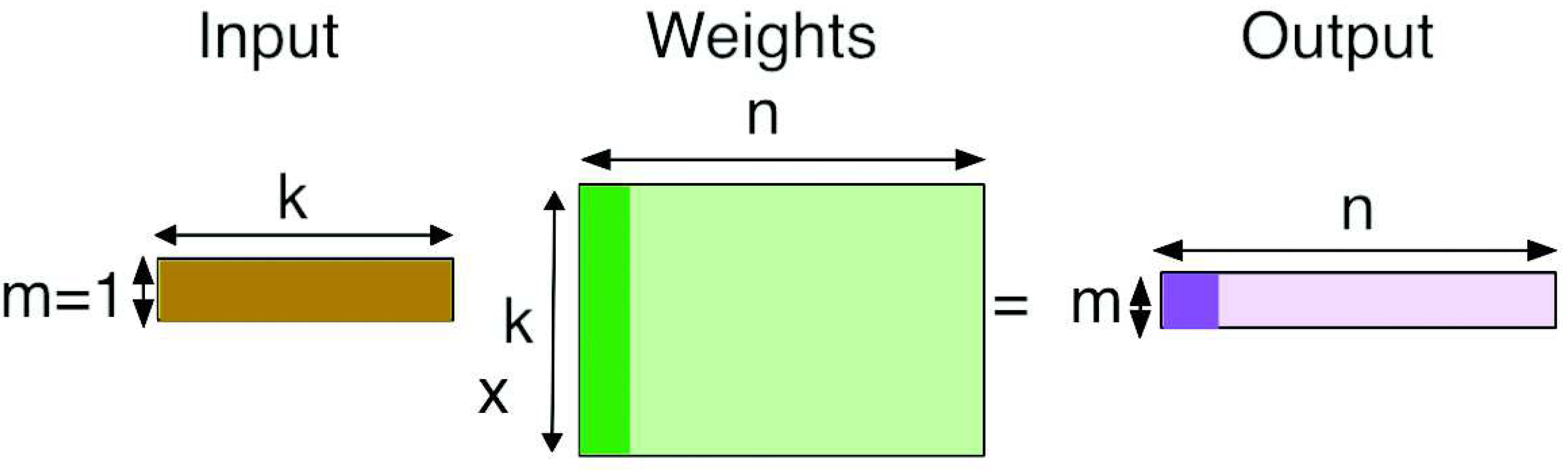}
    \caption{Fully-Connected Layer}
    \label{fig:fc}
    \end{subfigure}
    \hspace{0.2in}
    \begin{subfigure}[b]{0.22\textwidth}
    \includegraphics[width=\linewidth]{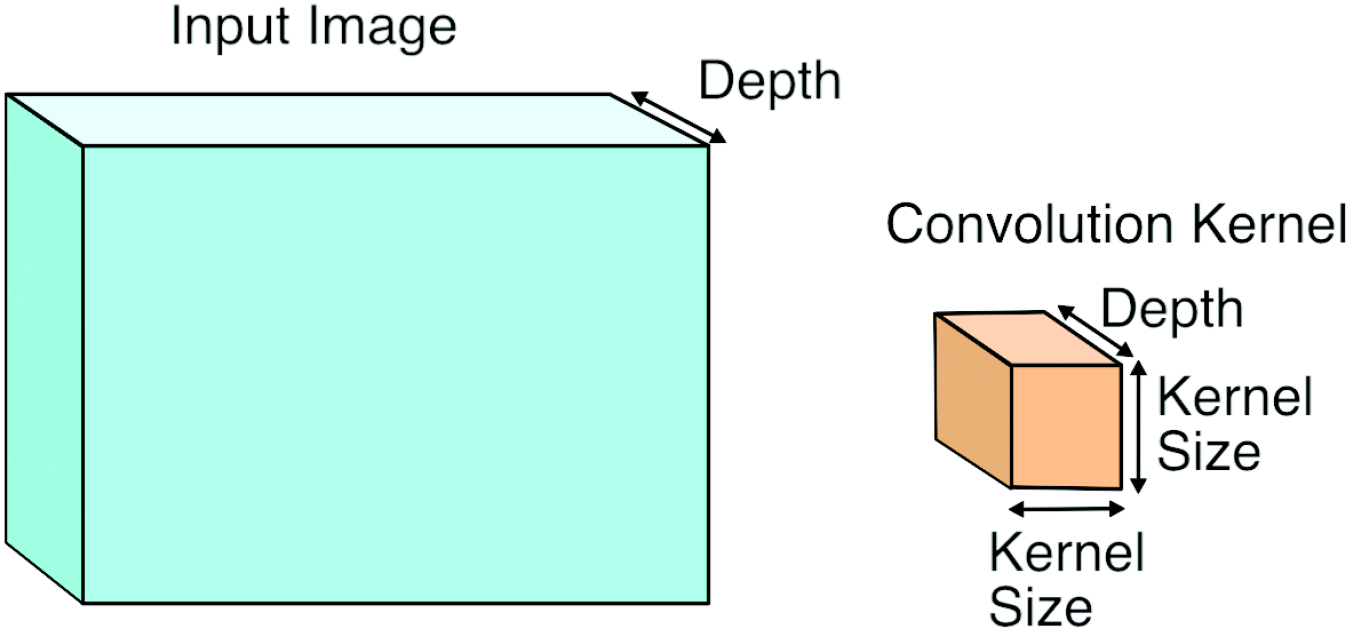}
    \caption{Convolutional Layer}
    \label{fig:cnn}
    \end{subfigure}
    \caption{Illustration of deep neural network layers}
    \label{fig:layers}
\end{figure}

Matrix multiplication is a critical operation in many machine learning algorithms, particularly in the domain of deep learning. Training parameters (weights) of a deep neural network in a vectorized fashion essentially involves multiplication of matrices with various sizes. 

Fully-Connected (FC) layers (Fig.~\ref{fig:fc}) and convolutional (Conv) layers (Fig.~\ref{fig:cnn}) are building blocks of feed-forward and convolutional neural networks~\cite{Warden15}. It is straightforward to identify matrix multiplication in computing output value of a FC layer: each input has $k$ elements, and FC layer has $n$ neurons each with $k$ weights. An FC layer is the multiplication of a $m \times k$ matrix ($m$ is sample size) and a $k\times n$ matrix. A Conv layer appears to be a specialized operation, but it can be computed with matrix multiplication after rearranging data in a matrix format: each depth-wise (channel) slice of input can be added into an input matrix as a row; similarly each kernel can be added into a kernel matrix as a column. Convolution operation becomes multiplication of those two matrices. When using AlexNet on image classification with ImageNet dataset, vast majority of computation time on forward pass (94.7\% on GPU, and 88.7\% on CPU) is consumed by Conv and FC layers~\cite{Jia14}.


\subsection{GEMM and Matrix Tiling}

\begin{figure}[h]
    \centering
    \includegraphics[width=2.0in]{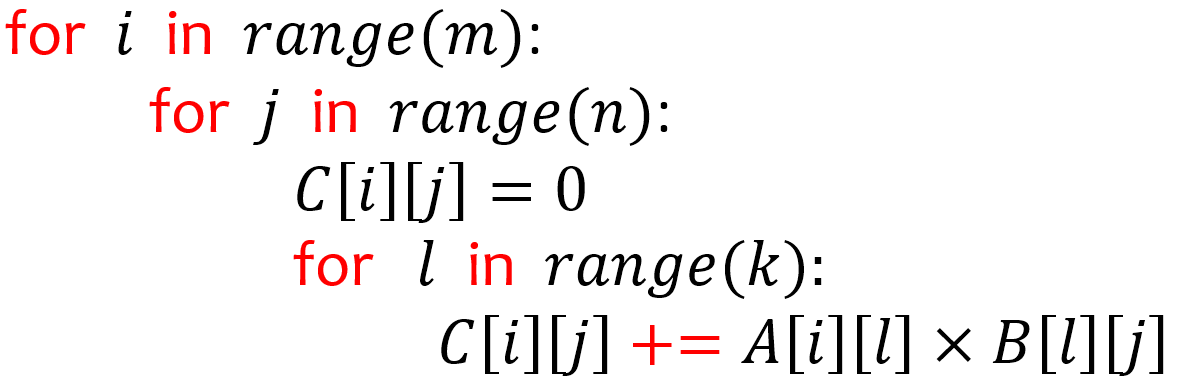}
    \vspace{-0.1in}
    \caption{Computing matrix multiplication}
    \label{fig:three}
\end{figure}

GEMM is a general procedure ubiquitously used in linear algebra, machine learning, statistics, and many other areas and is implemented in the BLAS (Basic Linear Algebra Subprograms) library \cite{BLAS2002}. It multiplies two input matrices to produce an output matrix. The key difference between GEMM in deep learning and regular matrix multiplication is that the input matrices handled in deep learning are normally much larger. For example, a single layer in a typical convolution neural network may require multiplication of a $256 \times 1024$ matrix by a $1024\times 128$ matrix to produce a $256 \times 128$ matrix. Regular three-for-loop (Fig.~\ref{fig:three}) computation requires 34 million ($256 \times 1024 \times 128$) floating point operations (FLOPs). Modern deep neural networks may have hundreds of convolutional layers (e.g. ResNet152~\cite{He15}), and such networks may need several billions of FLOPs to finish operations in all layers for an input image. 

\begin{figure}[t]
  \vspace{-0.5in}
    \centering
    \includegraphics[width=3.3in]{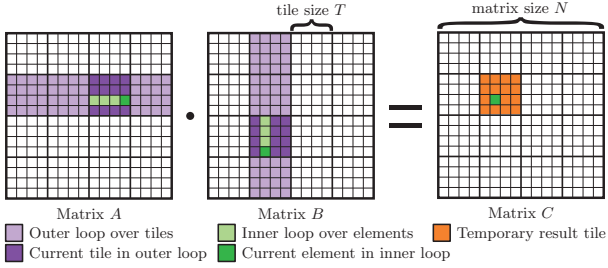}
    \vspace{-0.5in}
    \caption{An Example of Tiling Strategy}
    \label{fig:tiling}
\end{figure}

The time it takes to complete a GEMM computation largely depends on the cache hit rate of memory access. The large sizes of matrices usually forbid the entire matrices being loaded into memory or cache, however, GEMM can optimize memory access by iteratively splitting computation into smaller tiles, often referred to as the \emph{tiling process}.
A resulted matrix is initialized with zeros. GEMM uses outer products to compute part of a tile of the result and accumulates it on top of what has been stored in that tile. A tile is loaded from memory into cache and accumulates a new result on top of that. Fig.~\ref{fig:tiling}~\cite{Matthes17} illustrates a tiling strategy of GEMM.
 
Original memory access patterns need to be transformed to adapt to the cache policy of a particular hardware. It is not straightforward to decide an optimal tiling strategy because it requires accurate estimate of accessed array regions in loops to match with cache size of target hardware and meet other constraints. An optimal tiling configuration chooses a tile size for each loop to collectively achieve the lowest running time on target hardware.
\subsection{Problem Formulation}
\begin{figure}[htb]
    \centering
    \vspace{0.5cm}
    \includegraphics[width=3.3in]{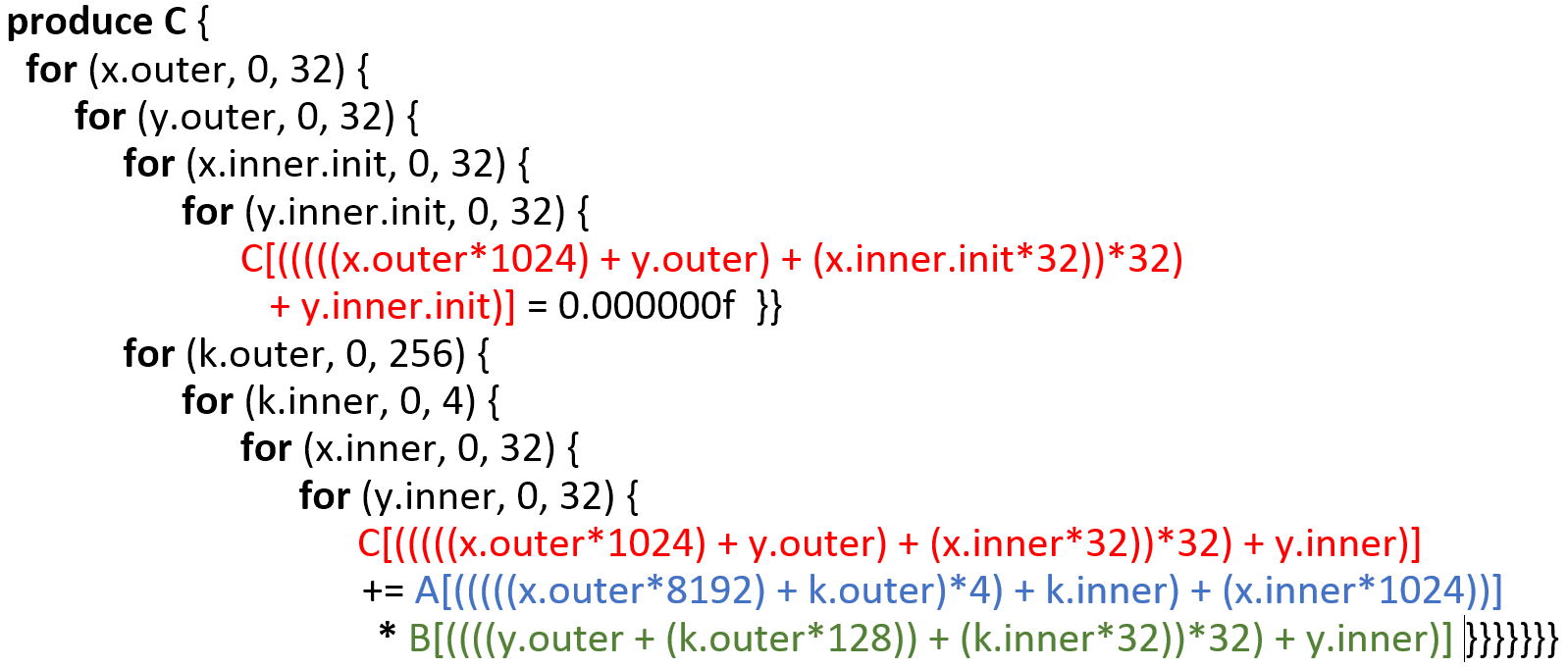}
    \caption{IR of GEMM with a blocking factor of 32}
    \label{fig:gemm_ir}
\end{figure}

We use TVM~\cite{Chen18} to investigate the performance of matrix tiling for GEMM. TVM facilitates tiling optimization by generating Intermediate Representation (IR) of a particular configuration. Fig.~\ref{fig:gemm_ir} is a simple example IR of GEMM tiling configuration with a blocking factor of 32 on x86 CPU for GEMM with $(m=1024, k=1024, n=1024)$ (short as $(1024,1024,1024)$).

\noindent
\textbf{Definition:}
Generally, a GEMM tiling configuration can be defined as 

\begin{equation}
    \overrightarrow{\xi} = \overrightarrow{\xi_m} \times \overrightarrow{\xi_k} \times \overrightarrow{\xi_n},
\end{equation}
where
\begin{equation}\label{eqn:m}
     \overrightarrow{\xi_m} =\{ \left[m_0,\ldots,m_i, \ldots m_{d_m-1} \right] |  \Pi_{i=0}^{d_m-1} m_i=m \},
\end{equation}
\begin{equation}\label{eqn:k}
     \overrightarrow{\xi_k} =\{ \left[k_0,\ldots,k_l, \ldots k_{d_k-1} \right] |  \Pi_{l=0}^{d_k-1} k_l=k \},
\end{equation}
\begin{equation}\label{eqn:n}
     \overrightarrow{\xi_n} =\{ \left[n_0,\ldots,n_j, \ldots n_{d_n-1} \right] |  \Pi_{j=0}^{d_n-1} n_j=n \}.
\end{equation}

Multiplication of two matrices $A(m\times k)$ and $B(k\times n)$ produces matrix $C(m\times n)$. $d_m$, $d_k$ and $d_n$ are the number of nested loops for each dimension $m$, $k$ and $n$, respectively.  $m_i, k_l, n_j$, $\forall i \in [0, d_m)$ $\forall l \in [0, d_k)$ $\forall j \in [0, d_n)$, are the number of iterations of a respective loop. The configuration in Fig.~\ref{fig:gemm_ir} is $m_0=m_1=32$, $k_0=256, k_1=4$, $n_0=n_1=32$, and $d_m=d_k=d_n=2$.

We can formulate the optimal tiling configuration search problem into the following optimization problem:

\begin{equation}\nonumber
    \begin{array}{l}
       \mathop {\min }\limits_{s} ~~{cost}(s ; m,k,n,d_m,d_k,d_n).
    \end{array}
\end{equation}

The objective is to find an optimal tiling configuration that has the minimal running time on target hardware. ${cost}$ denotes the running time for the configuration $s$, given the dimension of matrices ($m,k,n$) and the number of the nested loops on each dimension $d_m$, $d_k$, and $d_n$. 


\section{Methodology}\label{sec_method}
In solving the formulated GEMM problem, the state-of-the-art XGBoost tuner \cite{Tianqi2018} has been shown to outperform the other classic tuners including random search and genetic algorithm based search. Nevertheless, training the XGBoost model for a large configuration space would incur a high cost. In this section, we propose two new tuning methods and a new configuration search model which allows exploitation of relations between similar configurations.

\subsection{Configuration Search Modeling}

We model the configuration tuning problem as a Markov Decision Process (MDP), where each configuration is regarded as a unique state. We define a state as follows.

\begin{equation}
    s = \left[ s_m, s_k, s_n, J \right],
\end{equation}
where $s_m = \left[ m_0, m_1, \ldots, m_{d_m-1} \right] \in \xi_m$, $s_k = \left[ k_0, k_1, \ldots, k_{d_k-1} \right] \in \xi_k$, $s_n = \left[ n_0, n_1, \ldots, n_{d_n-1} \right] \in \xi_n$, and $J$ is a binary number indicating whether the state is legitimate.\footnote{The state configuration satisfies the conditions of Eqns.~\ref{eqn:m}-\ref{eqn:n}, and the numbers must be positive integers. Other constraints can be crafted to limit the search space and accelerate the search.}

As in the GEMM application, with similar configuration settings, i.e., the configuration parameters for each dimension of two states are equal or close, the performance of thw two states is more likely to be similar. Taking advantage of the relationship between similar configurations, and considering the constraints of the matrix size in each configuration. We define the action space as follows,
\begin{equation}
    \mathcal{A} = \emph{\{} {s_x[i] \leftarrow 2s_x[i] ~~\text{and}~~ s_x[j] \leftarrow s_x[j]/2 } \emph{\}},
\end{equation}
where $ \forall x \in \{m,k,n\}, \forall i, j \in [0, d_x),~\text{and}~ i \neq j $.

Accordingly, we define a step function $step$ as the transition from one state to another, 
\begin{equation}\label{fun:stepfun}
    s'= step(s,a).
\end{equation}
With the input of a particular action $a \in \mathcal{A}$, the current state $s$ transitions to state $s'$. 

\begin{figure*}[t!]
\vspace{0.5cm}
\centering
\includegraphics[scale=0.46]{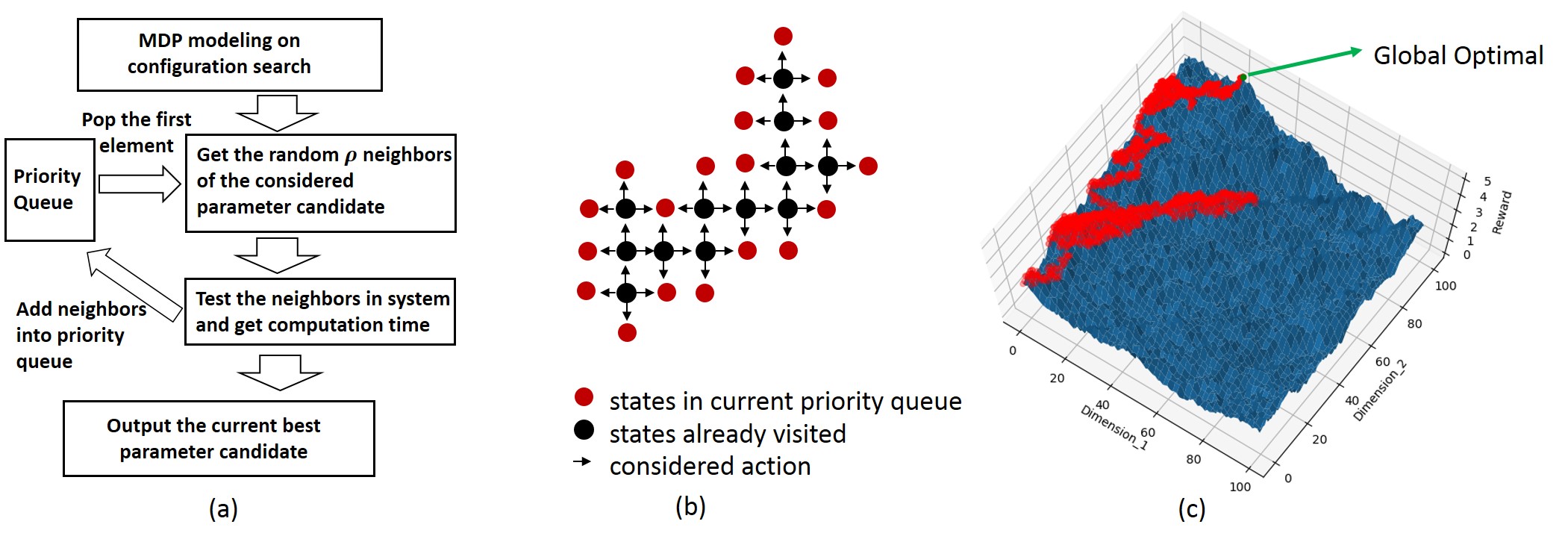}
\caption{G-BFS Method: (a) Flow chart; (b) Illustration; (c) Sample search trajectory}
\label{fig:gbfs}
\end{figure*}

In addition, 
if the agent takes action $a$ and transitions from state $s$ to state $s'$, we define the reward function as follows,

\begin{equation}
    r(s,a) = \frac{1}{{cost}(s'; m,k,n,d_m,d_k,d_n) }.
\end{equation}

Following MDP model, the agent is expected to determine its policy $\boldsymbol{\pi}$ so as to efficiently approach and discover a state $s^*$ with the lowest running time in the target hardware system. 

In the following subsections, we will propose two different configuration tuning approaches based on the configuration search model, guided by G-BFS and N-A2C reinforcement learning, respectively, followed by a discussion of their strengths in different scenarios.

\subsection{G-BFS Method}

The G-BFS method is guided by Greedy Best-First-Search and follows the flowchart in Fig. \ref{fig:gbfs}(a). We initialize an empty priority queue $\mathcal{Q}$ (ordered by increasing cost), an empty list $S_v$ to record all visited states, and a random or hand-crafted starting state $s_0$. We first test (i.e., run the configuration on target hardware) and enque the starting state $s_0$ and record its running time ${cost}(s_0)$ into the priority queue $\mathcal{Q}$. In each iteration, we deque the top configuration candidate $s$ from $\mathcal{Q}$, iterate through all actions $a\in \mathcal{A}$, and collect all corresponding neighbor states as

\begin{equation}
    g(s) = [s'=step(s,a) ~~ \forall a \in \mathcal{A}].
\end{equation}

We randomly select $\rho$ ($\rho \in \{1,2,\ldots,len(g(s)) \}$) states from $g(s)$, and test them in hardware. For each state $s'$ from $g(s)$, if $s'$ is legitimate and has not been visited before, we enque $s'$ and its running time ${cost}(s')$ into $\mathcal{Q}$ and add state $s'$ in the visited list $S_v$. If its running time ${cost}(s')$ is smaller than the current minimum running time, we set state $s'$ as the optimal state visited and record its running time as ${cost}_{min}$. The iteration continues until the priority queue is empty or the search time reaches the maximum time $T_{max}$ specified by the user. The current optimal state $s^*$ and its running time ${cost}_{min}$ are returned as tuning results. The summary of the algorithm is shown in Algorithm \ref{alg:gbfs}.

\begin{algorithm}[t!]
\caption{G-BFS Method}
\label{alg:gbfs}
\hrule
    \begin{algorithmic}[1]
    \vspace{.2cm}
    \STATE Initialization: $\mathcal{Q}$=PriorityQueue(), $S_v$, $s_0$
    \STATE $\mathcal{Q}$.push($({cost}(s_0), s_0)$);\\
    \STATE Add $s_0$ in $S_v$;\\
    \WHILE{$\mathcal{Q}\neq\O$ and $t_{search} < T_{max}$}
        \STATE $({cost}(s), s)$ = $\mathcal{Q}$.pop(); \\
        \STATE $\mathcal{B}$ = Take $\rho$ neighbors randomly from $g(s)$; \\
        \FOR{$s'$ in $\mathcal{B}$}
            \IF{$s'$ is legitimate and $s' \not\in S_v$}
                \STATE $\mathcal{Q}$.push($({cost}(s'), s')$); \\
                \STATE Add $s'$ in $S_v$;\\
                \IF{${cost}_{min} > {cost}(s')$}
                    \STATE ${cost}_{min} = {cost}(s')$; \\
                    \STATE $s^* = s'$;
                \ENDIF
            \ENDIF
        \ENDFOR
    \ENDWHILE
    \STATE Return: Configuration $s^*$ with ${cost}_{min}$.
    \end{algorithmic}
\hrule
\end{algorithm}

In Fig. \ref{fig:gbfs}(b), we illustrate the exploration in the middle of the tuning process, where the red nodes denote the state currently stored in the priority queue and the grey nodes are the visited states. In subsequent iterations, the method will explore from the $\rho$ most promising red nodes. In Fig. \ref{fig:gbfs}(c), we depict an example of 2-dimensional configuration search with a randomly generated reward function. The proposed G-BFS method is able to correct itself from exploring wrong directions and efficiently expand its neighborhood to the optimal states. Moreover, when $\rho = len(g(s))$, given unlimited tuning time, the algorithm is guaranteed to visit all configuration states.

\subsection{N-A2C Method}

\begin{figure*}[t!]
\centering
\vspace{0.5cm}
\includegraphics[scale=0.46]{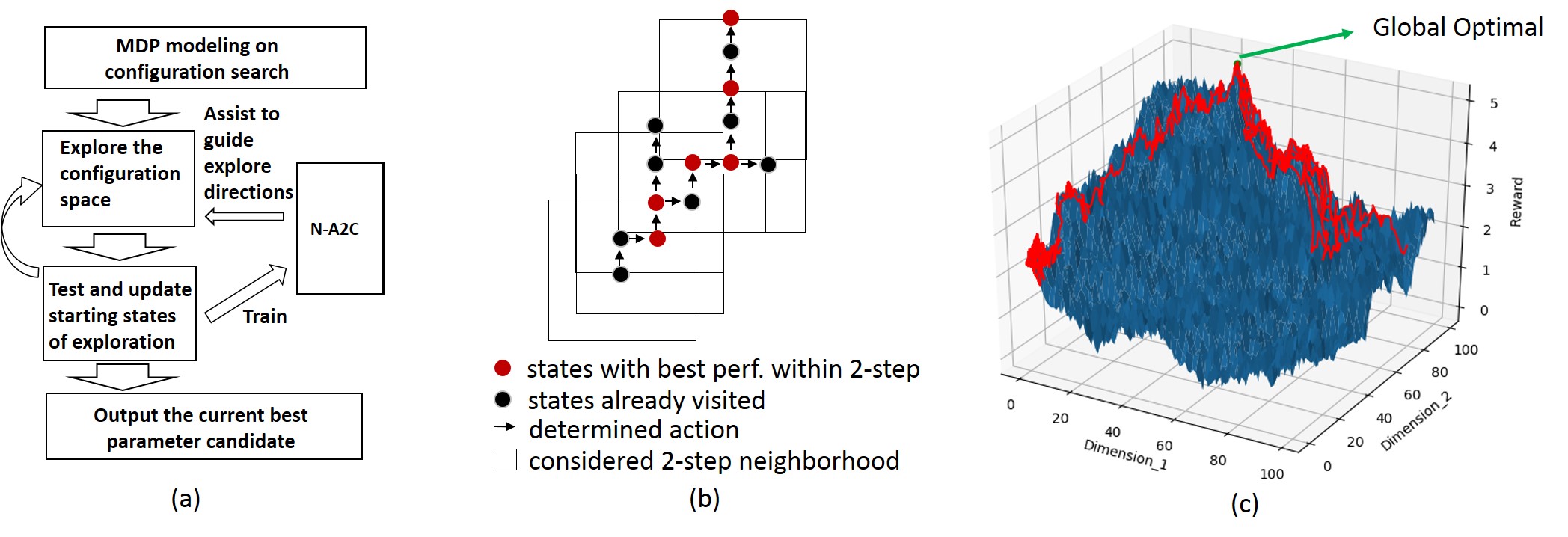}
\caption{N-A2C Method: (a) Flow chart; (b) Illustration; (c) Sample search trajectory}
\label{fig:r_a2c}
\end{figure*}

As the G-BFS method explores only one step from the considered state for each iteration, its performance may be affected when the cost from similar states exhibits large random noise. In the N-A2C method, as shown in Fig. \ref{fig:r_a2c}(a), for each episode, we explore in a $\varsigma$-step neighborhood, and the direction of exploration is guided by the A2C method \cite{bhatnagar2009natural}. The center of the exploration neighborhood is periodically updated with the optimal states ever visited.

We summarize the N-A2C method in Algorithm~\ref{alg:r_a2c}. We initialize a random or hand-crafted starting state $s_0$, a fixed-size memory buffer $\mathcal{M}$ to record the latest search information, and an empty hashtable $H_v$ to record all visited states with the associated cost. For the A2C model, both actor and critic initialize their neural networks with random weights. In each episode, from the same starting point, the agent explores $\mathcal{T}$ continuous steps. For each step, the agent follows the $\epsilon$-greedy algorithm, where with probability of $\epsilon$, the agent takes action $a$ guided by the policy $\pi(s)$ generated by the actor's neural network; and with probability of $1-\epsilon$, the agent chooses a random action $a$ from the current state. Based on the current state $s$ and action $a$, we get the next state $s'$ from Eqn.~\ref{fun:stepfun}. If the next state $s'$ has not been visited before, we add $s'$ into collected candidate set $\mathcal{B}_{collect}$. The process iterates until the number of collected states reaches the predefined threshold, and the hardware executes the GEMM computation code generated with the collected configuration candidates. 
The hashmap $H_v$ and memory buffer $\mathcal{M}$ are then updated with the configuration candidates and the associated running time. 
The exploration data stored in $\mathcal{M}$ is used to incrementally-train the neural networks of A2C.

\begin{algorithm}[htb!]
\caption{N-A2C Method}
\label{alg:r_a2c}
\hrule
    \begin{algorithmic}[1]
    \vspace{.2cm}
    \STATE Initialization: {$s_0$, $\mathcal{M}$, $H_v$, ${cost}_{min}$}
    \FOR{each episode}
        \WHILE{$len(\mathcal{B}_{collect})< len(\mathcal{B}_{test})$ }
            \STATE$s = s_0$; \\
            \FOR{each step until $\mathcal{T}$ steps}
                \IF{$rand()<\epsilon$}
                    \STATE $a$ follows $\pi(s)$;
                \ELSE
                    \STATE $a$ is randomly selected from $\mathcal{A}$;
                \ENDIF
                \STATE $s'=step(s,a)$; \\
                \IF {$s'$ not in $H_v$}
                    \STATE Add $s'$ in $\mathcal{B}_{collect}$;
                \ENDIF
                \STATE $s = s'$;
            \ENDFOR
        \ENDWHILE
        \FOR{$s'$ in $\mathcal{B}_{collect}$}   
            \IF{${cost}_{min} > {cost}(s')$}
                \STATE ${cost}_{min} = {cost}(s')$; \\
                \STATE $s^* = s'$; \\
                \STATE $s_0=s^*$; \\
            \ENDIF
            \STATE $H_v[s'] = {cost}(s')$; \\
            \STATE Store $(s, a, r(s,a), s')$ to $\mathcal{M}$, where $\forall s$, $\forall a$ satisfying $step(s,a) = s'$; \\
            \STATE Train actor's and critic's neural networks with $\mathcal{M}$;
        \ENDFOR
    \ENDFOR
    \STATE Return: Configuration $s^*$ with ${cost}_{min}$.
    \end{algorithmic}
    \hrule
\end{algorithm}

Generally, the proposed N-A2C method is able to efficiently search the optimal GEMM configuration with fixed exploration steps in each episode. Nevertheless, heuristics can be applied. Just like learning-rate decay in training deep neural networks, the exploration step $\mathcal{T}$ can have a decay process, i.e.,  starting with a large value and gradually reducing to a small number. In addition, the exploration step $\mathcal{T}$ can also increase to explore new configuration neighborhoods.

In Fig. \ref{fig:r_a2c}(b), we depict a simple exploration map with the proposed N-A2C method and $\mathcal{T} = 2$. Unlike Fig.~\ref{fig:gbfs}(b), the exploration neighborhood is defined as two steps from current states. In Fig. \ref{fig:r_a2c}(c), we show an example of 2-dimensional configuration with a randomly generated reward function. Due to the large randomness in the example, we set the exploration step $\mathcal{T}$ as 100 and the global optimal state is efficiently discovered with the guidance of the A2C algorithm.

\section{Experimental Results}\label{sec_simulation}
\begin{figure}[t!]
    \centering
    \begin{subfigure}[b]{0.48\textwidth}
    \includegraphics[width=\linewidth]{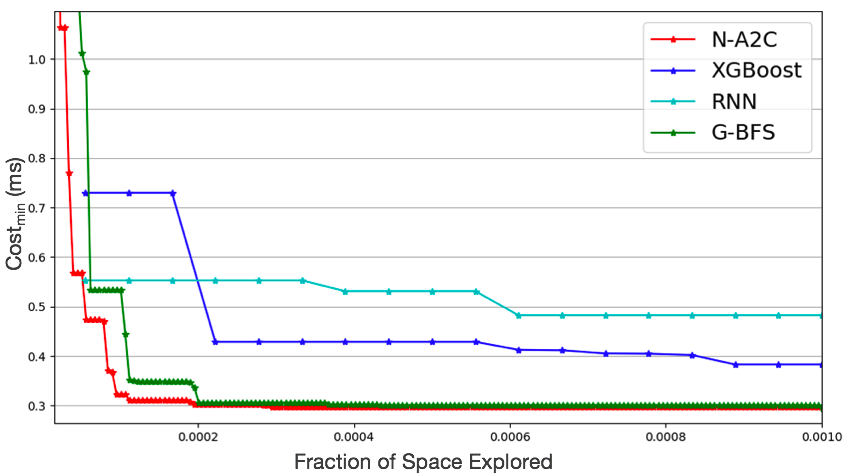}
    \caption{Optimal running time vs. fraction of explored configuration space}
    \label{fig:simu11}
    \end{subfigure}
    \hspace{0.2in}
    \begin{subfigure}[b]{0.48\textwidth}
    \vspace{0.2in}
    \includegraphics[width=\linewidth]{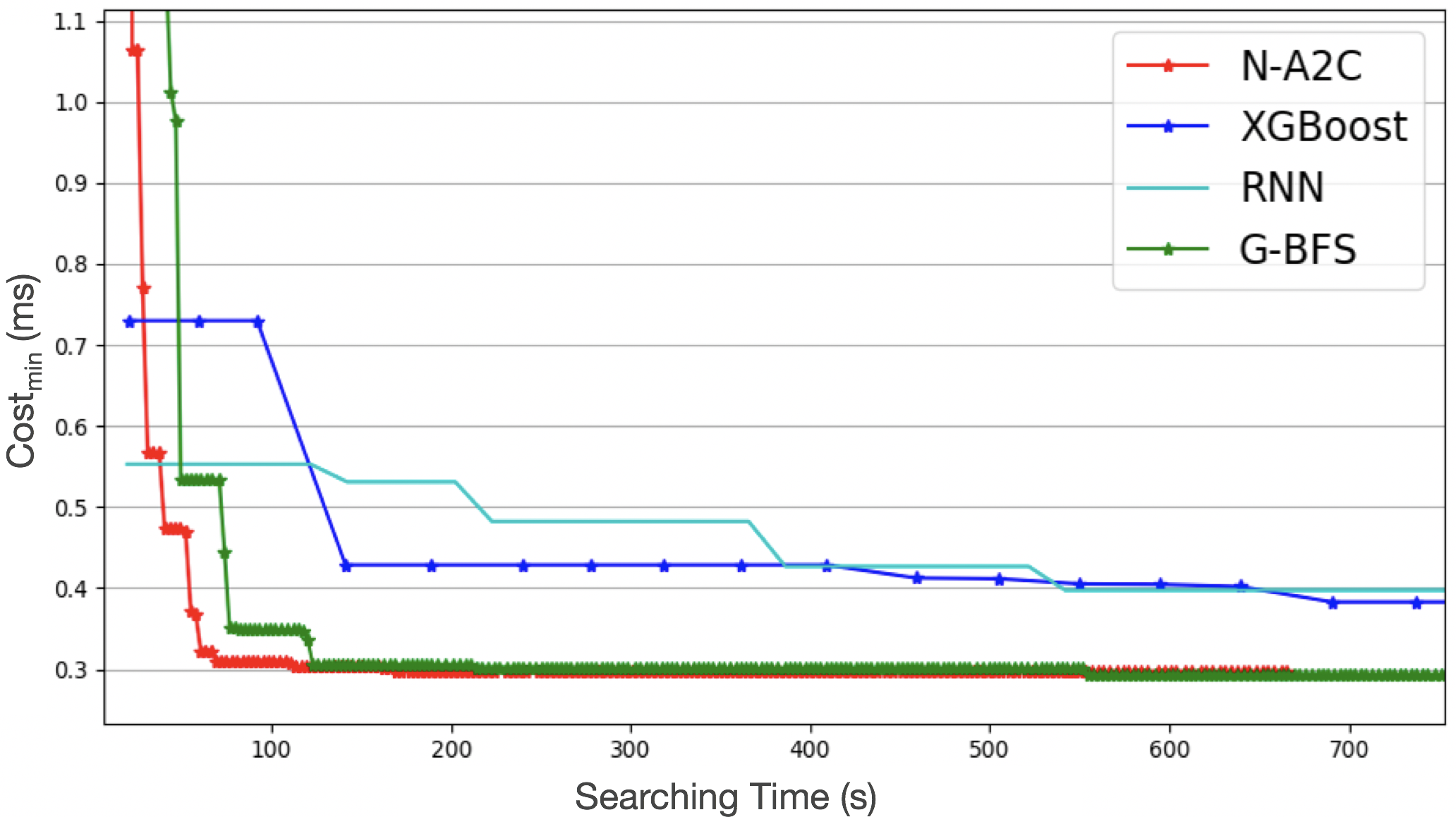}
    \caption{Optimal running time vs. searching time}
    \label{fig:simu12}
    \end{subfigure}
    \caption{Comparison of GEMM tuning for (1024, 1024, 1024)}
    \label{fig:1024}
\end{figure}

We evaluate the performance of the proposed GEMM configuration tuning approaches on Nvidia Titan Xp GPU in the TVM framework. Following similar settings in TVM for GPUs, we set the number of nested loops for each dimension as $d_m=4, d_k=2, d_n=4$. We set the random selection parameter $\rho=5$ for the G-BFS method, and the maximum number of search steps $\mathcal{T} =3$ for the N-A2C method. Without losing generality, we set the initial state for both methods as $s_0 = [[m,1,1,1],[k,1],[n,1,1,1]]$, which represents the configuration without multi-level matrix tiling. The performance of the proposed methods can be further improved by setting the initial state to a more meaningful configuration. In order to investigate the performance of the proposed methods, we compare them with state-of-the-art algorithms including the XGBoost method \cite{Tianqi2018} in TVM framework and the general configuration optimization method using a RNN controller by Google researchers. Unless otherwise specified, the computation time for each configuration is the arithmetic mean for 10 repeated trials on the tested GPU hardware.


Without losing their applicability to deep learning, we evaluate the four approaches on a perceptron network, which is the fundamental building block for state-of-the-art neural network architectures while mainly including GEMM for computation. We denotes the computation in the perceptron network as $\mathbf{Y}=\mathbf{W^T}\mathbf{X}$, where $\mathbf{W}\in \mathbb{R}^{(k,m)}, \mathbf{X}\in \mathbb{R}^{(k,n)}$, and $\mathbf{Y}\in \mathbb{R}^{(m, n)}$; $n$ is the batch size, $k$ is the input dimension, and $m$ is the output dimension. Accordingly, we denote the size of perceptron network in the format of $(m,k,n)$, which corresponds to matrices $A(m \times k)$ and $B(k \times n)$ for GEMM.


In Fig. \ref{fig:1024}, we set input dimension as $1024$, batch size as $1024$ and output dimension as $1024$. We first evaluate the optimal computation time discovered with respect to the fraction of visited configurations in Fig. \ref{fig:simu11}. Based on the sizes of matrices and the number of nested loops, there are $899756$ configuration candidates. With increasing fraction of visited configurations, the optimal cost in terms of hardware computation time discovered generally decreases. Compared with the XGBoost and RNN methods, the proposed N-A2C and G-BFS methods are able to discover the better configurations with lower fraction of visited configuration candidates. Fig. \ref{fig:simu12} plots the optimal cost discovered by four methods over time. It generally takes the proposed N-A2C and G-BFS methods less time to find the configuration with lower cost, compared with the XGBoost and RNN methods.

\begin{figure}[t]
    \centering
    \begin{subfigure}[b]{0.48\textwidth}
    \includegraphics[width=\linewidth]{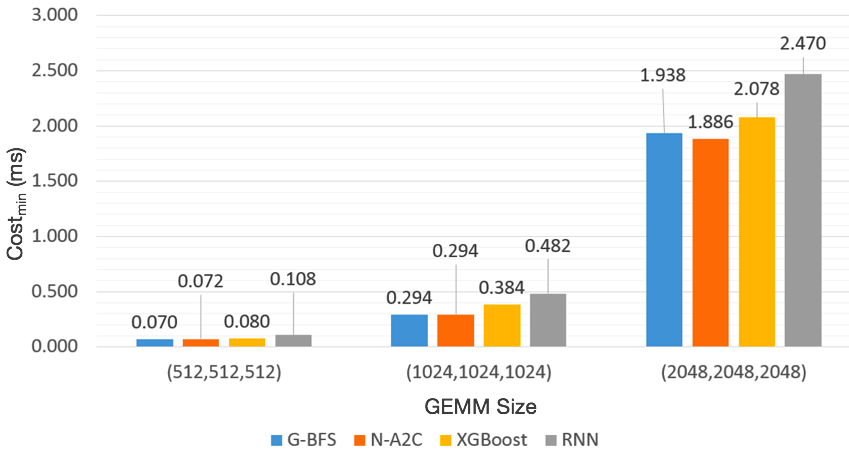}
    \caption{Visited configurations = 0.1\%}
    \label{fig:simu21}
    \end{subfigure}
    \hspace{0.3in}
    \begin{subfigure}[b]{0.48\textwidth}
    \vspace{0.2in}
    \includegraphics[width=\linewidth]{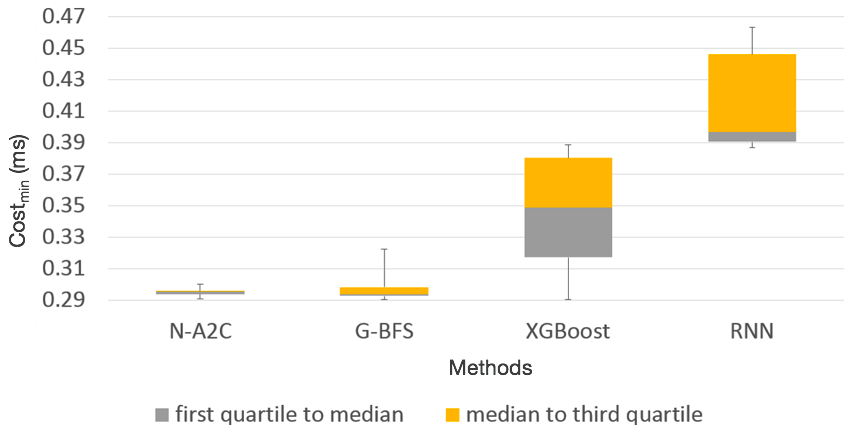}
    \caption{Searching time = 750 seconds}
    \label{fig:simu23}
    \end{subfigure}
    \caption{Performance comparison of tuners}
    \label{fig:multi_case}
    \vspace{-0.1in}
\end{figure}

In Fig. \ref{fig:multi_case}, we evaluate and compare the configuration tuning efficiency in the perceptron network. In Fig. \ref{fig:simu21}, we compare the discovered optimal computation time when the fraction of visited configuration candidates reaches 0.1\%. The total numbers of configuration candidates for $(512, 512, 512)$, $(1024, 1024, 1024)$, $(2048, 2048, 2048)$ matrix tiling are $484000$, $899756$, and $1589952$, respectively. As the sizes of matrices increase, longer computation time is required, and G-BFS and N-A2C can search configuration more efficiently than the XGBoost and RNN methods. Specifically, with 0.1\% exploration of $(1024, 1024, 1024)$'s configuration space, the proposed G-BFS and N-A2C methods are able to discover configurations of 24\% lower cost (computation time) than what the XGBoost method can find and configurations of 40\% lower cost than what the RNN method can find. The N-A2C method outperform the G-BFS method for larger matrix sizes (e.g. $(2048, 2048, 2048)$), as the N-A2C method is able to go multiple steps from the current state. In Fig. \ref{fig:simu23}, we compare the cost of best configuration discovered when the tuning time is limited to 750 seconds. In order to show the variance of performance incurred by random exploration for each method, we use a box plot with the minimum, first quartile, median, mean, third quartile, and maximum values for 10 trials on the $(1024, 1024, 1024)$ tiling. Our methods not only achieve better mean and median results, but also exhibit more stable behaviors (less variance) than the other two methods.


\section{Conclusion}\label{sec_conclusion}
In this paper, we propose a Greedy Best First Search (G-BFS) method and a Neighborhood Actor Advantage Critic (N-A2C) method for compiler-level GEMM optimization, taking advantage of performance of neighborhood configurations. The G-BFS method, though being lightweight, outperforms the XBboost and RNN methods consistently; and the N-A2C method performs even better for large matrices. Empirical results show that both methods achieve significant performance improvement over  state-of-the-art tuning methods such as those using XGBoost and RNN controller. Both methods are general in the sense that they are applicable to other compiler-level tuning tasks and can be used for optimization of other tensor operators with large configuration space.

{
\begin{spacing}{0.92}
\footnotesize
\bibliographystyle{icml2019}
\bibliography{icml}
\end{spacing}
}

\end{document}